\documentclass[conference]{IEEEtran}
\IEEEoverridecommandlockouts

\usepackage{cite}
\usepackage{amsmath,amssymb,amsfonts}
\usepackage{algorithmic}
\usepackage{graphicx}
\usepackage{float}
\usepackage{textcomp}
\usepackage{multirow}
\usepackage{caption}
\usepackage{xcolor}

\def\BibTeX{{\rm B\kern-.05em{\sc i\kern-.025em b}\kern-.08em
    T\kern-.1667em\lower.7ex\hbox{E}\kern-.125emX}}
\begin{document}

\title{HyperSMOTE: A Hypergraph-based Oversampling Approach for Imbalanced Node Classifications\\
}

\author{
\IEEEauthorblockN{Ziming Zhao}
\IEEEauthorblockA{\small{\textit{School of Information}} \\
\textit{University of Michigan, USA}\\
zhziming@umich.edu
}
\and
\IEEEauthorblockN{Tiehua Zhang$^{\dagger}$}
\IEEEauthorblockA{\small{\textit{College of Electronics and Information Engineering}}\\
\textit{Tongji University, China}\\
tiehuaz@tongji.edu.cn}
\and
\IEEEauthorblockN{Zijian Yi and Zhishu Shen$^{\dagger}$}
\IEEEauthorblockA{\small{\textit{School of Computer Science and AI}}\\
\textit{Wuhan University of Technology, China}\\
yzjyjs@whut.edu.cn, z\_shen@ieee.org}


\thanks{$^{\dagger}$\textit{Corresponding author: Tiehua Zhang and  Zhishu Shen}}

}

\maketitle

\begin{abstract}

Hypergraphs are increasingly utilized in both unimodal and multimodal data scenarios due to their superior ability to model and extract higher-order relationships among nodes, compared to traditional graphs. However, current hypergraph models are encountering challenges related to imbalanced data, as this imbalance can lead to biases in the model towards the more prevalent classes.
While the existing techniques, such as GraphSMOTE, have improved classification accuracy for minority samples in graph data, they still fall short when addressing the unique structure of hypergraphs. Inspired by SMOTE concept, we propose HyperSMOTE as a solution to alleviate the class imbalance issue in hypergraph learning. This method involves a two-step process:  initially synthesizing minority class nodes, followed by the nodes integration into the original hypergraph. We synthesize new nodes based on samples from minority classes and their neighbors. At the same time, in order to solve the problem on integrating the new node into the hypergraph, we train a decoder based on the original hypergraph incidence matrix to adaptively associate the augmented node to hyperedges. We conduct extensive evaluation on multiple single-modality datasets, such as Cora, Cora-CA and Citeseer, as well as multimodal conversation dataset MELD to verify the effectiveness of HyperSMOTE, showing an average performance gain of 3.38\% and 2.97\% on accuracy, respectively.


\end{abstract}


\section{Introduction}


Hypergraph learning has gained increasingly attention owing to its potential in modeling and extracting high-order correlations within data compared to the ordinary graphs~\cite{hypergraph}. A key benefit of the hypergraph is that a hypergraph can encompass various types of hyperedges and nodes, while one hyperedge can connect multiple nodes, enabling a more accurate representation of group concepts that aligns with the deep relationships among entities in certain contexts~\cite{zhang2023learning}. This characteristic is particularly relevant in areas such as recommendation systems in social media\cite{10.1145/3477495.3531868}, emotion detection in multimodal dialogue\cite{Research}, and multi-source sleep quality assessment\cite{SleepHGNN}. However, similar to the cases in ordinary graphs, hypergraphs also face significant challenges when dealing with class imbalances issues.



In machine learning field,  class imbalance occurs  when a dataset is skewed, resulting in one class significantly outnumbering the others. This imbalance can cause models that excel in identifying the dominant class while struggling with the underrepresented one. In general,  solutions for tackling imbalanced data classification can be classified into three categories: data-level models\cite{Over-Sampling,Under-sampling}, algorithm-level models\cite{cost-sensitive}, and hybrid models\cite{hybrid1,hybrid2}. Data-level models aim to achieve the balance by either oversampling the minority class or undersampling the majority class\cite{math12132064}. Algorithm-level models enhance the model's sensitivity to the classification loss of the minority class by adjusting the class loss weights\cite{al}. Hybrid models combine elements of both data-level and algorithm-level strategies\cite{SMOTEBoost}. One of the most classic models is SMOTE~\cite{smote}, which synthesizes new instances in non-Euclidean data by selecting the two closest samples from the minority class. GraphSMOTE~\cite{graphsmote} applies the principles of SMOTE to graph learning by introducing innovative schemes for creating connections between newly generated nodes and existing nodes in the graph structure. Nonetheless, GraphSMOTE is not applicable to accommodate the unique structure of hypergraphs. The objective of this paper is to explore a general solution for pairwise ordinary graphs to alleviate the class imbalance issue.

Our work includes two main steps: The first step involves generating new nodes from the minority class. In a hypergraph topology that lacks physical meaning, such as multi-modal dataset MELD~\cite{MELD}, the nearest neighbor is defined as the most similar minority class node. In contrast, for data with explicit graph structure like Cora~\cite{cora}, the nearest neighbor is determined by the average of all hyperedges associated with the node to simulate the context. In this paper, the existing nodes are referred to as target nodes, while the newly generated node is the augmented node. It is crucial to focus on the alignment among various modalities in multimodal data, as they are inter-related and can influence each other. As a solution, we propose designating one modality as the dominant mode for the above operations, with the other modalities being referred to as following modes, replicating the operations of the dominant mode.



Subsequently, it is essential to construct the integration of augmented nodes into the hypergraph. Four potential methodologies are employed for this construction:  \textbf{Nearest Source Node Assignment}: it involves assigning the new node to the hyperedge corresponding to the nearest source node, thereby minimizing spatial discrepancies. \textbf{Comprehensive Allocation:} the augmented node is allocated to all hyperedges that encompass the source nodes, facilitating a broader connectivity perspective. \textbf{Model-Based Hyperedge Selection:} it selects hyperedges based on the underlying physical significance of the model, ensuring that the integration is contextually relevant. \textbf{Adaptive Decoding Mechanism:} it involves training a decoder utilizing the original hypergraph incidence matrix to dynamically associate the augmented node with appropriate hyperedges. 

The main contributions are as follows: (1) We propose HyperSMOTE, a general oversampling approach designed for application in both unimodal and multimodal hypergraphs, as well as on graphs with and without physical meaning. This research extends the applicability of SMOTE from pairwise ordinary graph learning to the more complex realm of hypergraph learning, thereby enhancing its performance when encoding the imbalanced data. (2) We augment new node features via minority class samples encompassed within same hyperedge, rather than relying on the most similar nodes adopted in the prior SMOTE work, which in turn oversamples minority nodes to the balanced sample distributions in the dataset. (3) We rigorously validate the HyperSMOTE with several widely recognized datasets. The experimental results substantiate the effectiveness of our approach against state-of-the-art methods.

\section{Related Work}

\subsection{Hypergraph Learning}


Compared with the ordinary graphs that are limited to pairwise connections, hypergraphs represent an advanced form of graph learning that can capture high-order correlations within data~\cite{hypergraph}. By allowing multiple nodes to be connected through a single hyperedge, hyperedges can represent diverse types of relationships, thus mimicking real-world structures while preserving abundant information~\cite{HGNN}. To date, several variants of hypergraph learning have emerged, including HGNN~\cite{HGNN}, HGNN+~\cite{gao2022hgnn+}, HyperAttn~\cite{bai2021hypergraph}  and HyperGCN~\cite{yadati2019hypergcn}. Despite differences in implementation, these hypergraph models all employ a convolutional approach, where the information from connected nodes is aggregated via hyperedges and subsequently transmitted back to the nodes. Hypergraph learning has demonstrated its effectiveness in various data association tasks~\cite{XiaKDD22,Yan_2020_CVPR,Ruan2021ExploringCA}. Although there have been significant advancements in hypergraph learning for data modeling and multimodal fusion, the class imbalance issues remains unresolved, hindering the broader adoption of hypergraph techniques.


\subsection{Class Imbalance Problem}

Class imbalance is a critical issue in machine learning, which often leads to generated models perform well on the majority class while underperforming on the minority class~\cite{smote}. Such an imbalance can introduce model bias and hinder generalization, as the model may overlook important patterns within the minority class.

Various strategies, including resampling techniques such as oversampling the minority class~\cite{Over-Sampling} or undersampling the majority class\cite{Under-sampling}, are proposed to address this issue. Additionally, cost-sensitive learning\cite{cost-sensitive} can be used to impose greater penalties for misclassifying the minority class. Overall, tackling class imbalance is essential for developing robust machine learning models, and experimenting with different strategies is crucial for achieving balanced performance.

Despite the notable success of GraphSMOTE in extending the concept of SMOTE to graph learning through the incorporation of learnable edge predictors, it is not applicable to hypergraph learning due to three main reasons: \textbf{Structural Differences:} Hyperedges possess fundamentally different structures compared to ordinary graph edges.
\textbf{Contextual Representation:} In hypergraphs with practical significance, selecting the most similar node as the neighbor for generating a new node fails to accurately represent the contextual relationships of the node, leading to discrepancies with the actual node.
\textbf{Challenges in Multimodal Integration:} In multimodal hypergraphs, GraphSMOTE encounters difficulties in synthesizing multimodal information that closely resembles the original data. This challenge arises from the complex interactions and constraints among different modalities associated with the same entity.

\section{Methodology}
\begin{figure*}[tb!]
    \centering
    \includegraphics[width=1\linewidth,height = 0.28\textwidth]{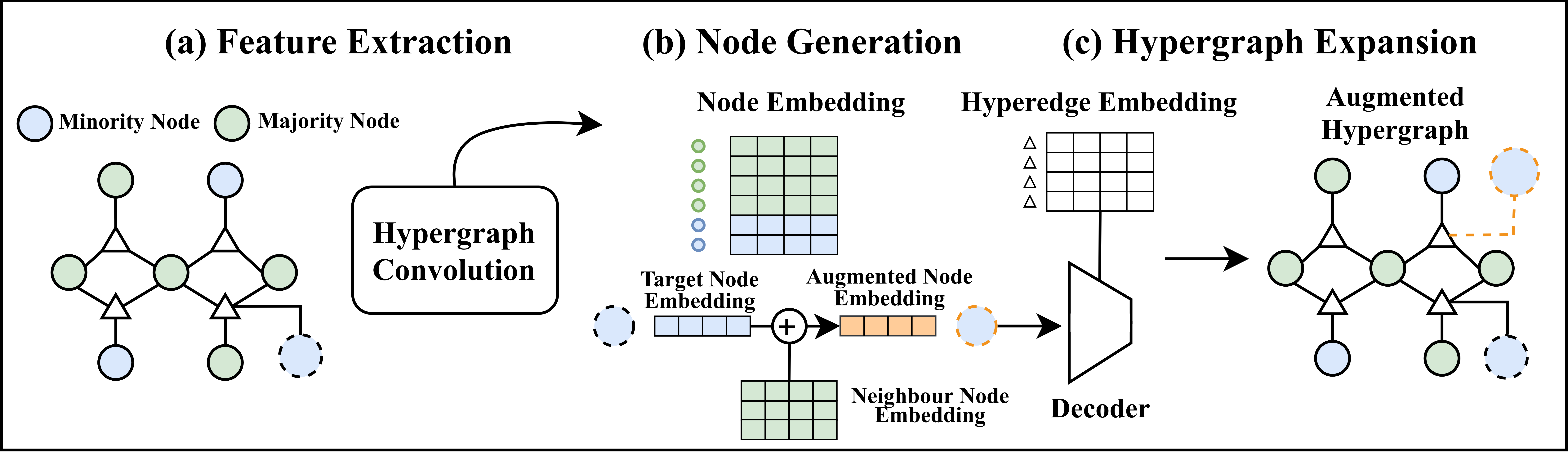}
  \caption{An overview of HyperSMOTE}
  \label{fig:all} 
  \vspace{-5mm}
\end{figure*} 
The overall pipeline of HyperSMOTE is composed of three modules, including the feature extraction, node feature generation, and hypergraph expansion. The pipeline of HyperSMOTE is shown in Fig~\ref{fig:all}. 
\subsection{Feature Extraction}
The initial hypergraph can be represented with notation $G = \{V, \mathcal{E}, X\}$, where $X\in R^{|V|\times D}$ denotes the raw feature of each node. Following the concept of message passing from MPNN\cite{gilmer2017neural}, convolution on hypergraph can be considered as a two-step process. Information from nodes are first aggregated to their corresponding edges, resulting in the feature embeddding of hyperedges. Then the message is propagated back to nodes, which further updates the node embeddings. The process of hypergraph convolution can be formulated as follows:
\begin{equation}
E_{\mathcal{E}} = \textbf{Aggr}(\mathcal{E}, \sigma (W_1X))
\end{equation}
\begin{equation}
E_{v} = \textbf{Aggr}(\mathcal{E}, \sigma(W_2E_\mathcal{E}))
\end{equation}
where $E_\mathcal{E}\in R^{|\mathcal{E}|\times D}$ stands for the hyperedge embedding and $E_v\in R^{|V|\times D}$ represents the updated node embedding. $W_1$ and $W2$ are linear projection matrices and $\sigma$ stands for the activation function. The aggregation is chosen with a mean pooling layer, which equivalently provides normalization based on the degree of hyperedges and nodes.

\subsection{Node Generation}
In order to augment the training samples from minority classes to alleviate the class imbalance problem by increasing the size of the training set, the augmentation process is repeated multiple times for each single sample within the minority classes. In Fig~\ref{fig:all}(a), samples from majority classes are represented with green nodes while samples from the minority classes are denoted in blue. The specific target node that is being augmented is represented by the blue node surrounded by black dash boundary, which is denoted as $v_t$, and the paired augmented node is represented by the blue node surrounded by orange dash boundary, which is denoted as $v_g$.
It is required to build up the feature embedding for $v_g$ from the context of hypergraph. To make sure the synthesized embedding lies in the same feature space with the target node $v_t$, the feature embeddings from both $v_t$ and its neighbors should be considered for feature generation. Denote the feature embedding of the augmented node as $E_{v_g}$, and it is generated as follows:
\begin{equation}
E_{v_g} = \tau E_{v_t} + (1-\tau)\textbf{Mean}(\{E_{v_i}| v_i\in N(v_t)\})
\end{equation}
where $N(v_g)$ represents the neighbor nodes of $v_t$ and a certain node is considered as the neighbor node of $v_t$ if it is connected by any hyperedge that includes $v_t$. A mean pooling is applied to extract the embeddings of all its neighbors. $\tau$ herein refers to the hyperparameter that determines the weight between the target node embedding and neighbor node embeddings. Based on the general assumption that nodes with close attributes are more likely to be connected, by adding the feature of the target node and its neighbor embeddings, it is possible to generate a synthesis node that is similar to the target node in the feature space. The generated embedding is illustrated as the orange vector in Fig~\ref{fig:all}(b). Moreover, the label of all augmented nodes should maintain the same with the target nodes, which means
$$
Y(v_g) = Y(v_t).
$$

\subsection{Hypergraph Expansion}
To fuse the augmented node into the hypergraph as well as perform hypergraph convolution on both original nodes and augmented nodes, the augmented nodes need to be adaptively appended to the inital hypergraph. A specific form of decoder is designed to reconstruct the hypergraph incidence matrix $H$. The decoder can be fomulated with the following transformation:
\begin{equation}
\hat{H}_{\epsilon, v_g} = \sigma(E_{v_g}\cdot P\cdot E_{\epsilon})
\end{equation}
where $P\in R^{D\times D}$ denotes the learnable projection matrix that projects the dot product between the feature embedding of hyperedges and nodes into a scalar. As all elements of hypergraph incidence matrix take on value of either 0 or 1, the sigmoid activation function is applied to limit the range of the decoder output. The decoder is trained based on the original incidence matrix with no augmented nodes. By optimizing the following loss
\begin{equation}
L_P = -\sum_{v_i\in V}\sum_{\epsilon \in \mathcal{E}}(H_{\epsilon, v_i}\log \hat{H}_{\epsilon, v_i} + (1-H_{\epsilon, v_i})\log(1-\hat{H}_{\epsilon, v_i})),
\end{equation}
the decoder is able to capture the topology of the hypergraph and reconstruct the hypergraph incidence matrix based on node embeddings and hyperedge embeddings. 
The augmented node will be attached to a single hyperedge that is most relevant to it, in other words, 
\begin{equation}
H_{\epsilon, v_g}  = 
\begin{cases}1, & \text{if } \hat{H}_{\epsilon, v_g} = \textbf{max}(\{\hat{H}_{\epsilon_i, v_g}, \epsilon_i\in \mathcal{E}\}) \\ 0, & \text { otherwise }\end{cases}
\end{equation}
which means the augmented node will be attached to the hyperedge with the highest inference probability from the incidence matrix predicted by the decoder. The extended hyperedge is illustrated as the orange dash line from Fig~\ref{fig:all}(c). The overall augmentation process of HyperSMOTE is completed before task-specific supervised training, which means the node augmentation and hyperedge expansion are performed to the dataset independently and a single round of augmentation can be generally applied to multiple  downstream tasks on the same dataset.

\section{Experiments}

\subsection{Experimental Setting}
HyperSMOTE is evaluated on three widely adopted datasets from graph learning field and a multimodal dataset from the field of emotional recognition. We apply HyperSMOTE on Cora, Cora-Cocitation and Citeseer datasets, where each hyperedge explicitly represents co-citation or co-authorship relationships between papers. While for multimodal dataset MELD, each hyperedge represents the correlation among audio, visual and textual modalities. We apply this setting of datasets in order to justify that HyperSMOTE is able to augment both hyperedges with practical significance and hyperedges with abstract usage for feature fusion.

We compare the proposed model of HyperSMOTE with six baseline methods for hypergraph learning on citation datasets. On multimodal dataset, we apply three baselines with the same backbones as those used in citation data, ensuring a fair comparison. \textbf{GCN}~\cite{kipf2016semi} performs graph convolution from a spectral prospective based on the normalized Laplacian Matrix. \textbf{GAT}~\cite{velivckovic2017graph} introduces attention mechanism into the graph convolution process. \textbf{GraphSAGE}~\cite{hamilton2017inductive} performs graph convolution from a spatial prospective by aggregating message from nodes. \textbf{HyperAttn}~\cite{bai2021hypergraph} applies attention mechanism on hypergraph convolution. \textbf{HGNN}~\cite{HGNN} applies spectral theory of hypergraphs and use the Laplacian Matrix for hypergraph convolution. \textbf{HGNN+}\cite{gao2022hgnn+} designs a two-stage hypergraph convolution based on message passing from spatial domain and trains parameters for individual hyperedge group. \textbf{MMGCN}\cite{wei2019mmgcn} applies GAT with attention weights representing modality relevance. \textbf{HAUCL}\cite{yi2024multimodal} uses HGNN as its backbone and some extra reconstruction modules to further enhance modality fusion.

Following the data splitting strategy in~\cite{gao2022hgnn+}, 140 samples are applied for training, 500 samples for validation and 1000 samples are applied for testing on Cora and Cora-CA datasets. On Citeseer, 120 samples are used for training, 500 used for validation and 1015 used for testing.  On MELD, training set contains 9939 samples, and validation set contains 1109 samples. Another 2610 samples are used for testing. For all datasets, we define three categories among all classes with the lowest number of labelled nodes in the whole dataset as minority category. By default, HyperSMOTE will repeat the augmentation to each training sample from minority category 3 times and it will not augment samples from majority categories. The default value of $\tau$ is set to 0.3.

\begin{table}[]
\centering
\captionsetup{justification=centering}
\caption{Experimental Results on Citation Datasets}
\resizebox{9cm}{!}{%
\begin{tabular}{lllllllll}
\hline \multirow{2}{*}{ Methods } & \multicolumn{2}{c}{ Cora } & & \multicolumn{2}{c}{ Cora-CA} & & \multicolumn{2}{c}{ Citeseer } \\
\cline { 2 - 3 } \cline { 5 - 6 } \cline { 8 - 9 } & Acc & Marco-F1 & & Acc & Marco-F1 & & Acc & Marco-F1 \\
\hline GCN & 0.763 & 0.749 & & 0.705 & 0.688 & & 0.638 & 0.597\\
GAT & 0.795 & 0.748 & & 0.711 & 0.697 & & 0.647 & 0.608\\
GraphSAGE & 0.732 & 0.716 & & 0.698 & 0.676 & & 0.593 & 0.557\\
HyperAttn & 0.762 & 0.749 & & 0.705 & 0.694 & & 0.639 & 0.628\\
HGNN & 0.771 & 0.759 & & 0.714 & 0.699 & & 0.640 & 0.626 \\
HGNN$^{+}$ & 0.767 & 0.755 & & 0.717 & 0.703 & & 0.664 & 0.649 \\
HyperSMOTE & 0.793 & 0.784 & & 0.726 & 0.715 & & 0.672 & 0.663\\
\hline
\end{tabular}
}
\label{citation}
\end{table}

\begin{table}[]
\centering
\captionsetup{justification=centering}
\caption{Experimental Results on Multimodal Dataset}
\resizebox{5cm}{!}{%
\begin{tabular}{lll}
\hline \multirow{2}{*}{ Methods } & \multicolumn{2}{c}{ MELD } \\
\cline { 2 - 3 } & Acc & Marco-F1 \\
\hline MMGCN(GAT) & 0.593 & 0.578 \\
HyperAttn & 0.675 & 0.652  \\
HAUCL(HGNN) & 0.680 & 0.667  \\
HyperSMOTE & 0.684 & 0.672\\
\hline
\end{tabular}
}
\label{multimodal}
 \vspace{-5mm}
\end{table}

\subsection{Experimental Result}
The experiment results are summarized in Table~\ref{citation} and Table ~\ref{multimodal}. Since we aim to evaluate models that can best improve the performance of minority categories while not impairing the performance of other categories, Macro-F1 score and Accuracy are applied as evaluation metrics. 

HyperSMOTE is able to reach an average performance gain of 3.84\% in Macro-F1 score and 3.38\% in Accuracy on citation datasets and 3.85\% in Macro-F1 score and 2.97\% in Accuracy on multimodal datasets. The results indicate that HyperSMOTE enhances model performance by balancing the number of training samples across all classes. It is achieved by generating in-distribution nodes that conform to the training data's distribution and adaptively attach the augmented node to the most relevant hyperedge, informed by the data's topology. Such augmentation of HyperSMOTE acts as a solution to imbalanced labels.

\begin{table}[H]
\centering
\captionsetup{justification=centering}
\caption{Sensitivity Analysis on \# of Augmented Nodes}
\begin{tabular}{c|c|c|c}
\hline \# of Augmented Nodes & Cora & Cora-CA & Citeseer \\
\hline 1-Sample & 0.771 & 0.720 & 0.665 \\
3-Sample & 0.789 & \textbf{0.726} & 0.669 \\
Adaptive-Sample & \textbf{0.793} & 0.725  & \textbf{0.672} \\
\hline
\end{tabular}
\label{number}
 \vspace{-3mm}
\end{table}

\begin{table}[H]
\centering
\captionsetup{justification=centering}
\caption{Variants of Hyperedge Expansion}
\begin{tabular}{c|c|c|c}
\hline Variants & Cora & Cora-CA & Citeseer \\
\hline Random & 0.757 & 0.703 & 0.636 \\
Closest Node & 0.791 & 0.720 & 0.667 \\
Closest Hyperedge & \textbf{0.793} & \textbf{0.726}  & \textbf{0.672} \\
\hline
\end{tabular}
\label{hyper-extension}
\end{table}

The performance gain of HyperSMOTE is mainly attributed to node feature generation and hyperedge expansion. We hereby give detailed analysis of HyperSMOTE on these two features based on citation datasets. We first show the performance change caused by number of augmented nodes in Table~\ref{number}. The model performs the best when the number of augmented nodes is adaptively calculated based on the ratio between training samples from majority classes and minority classes. In other words, the more imbalanced the dataset is, the more augmentation is required to alleviate the issue. For hyperedge expansion, we design three variants for comparison in Table~\ref{hyper-extension}. When the augmented node is randomly added into an existing hyperedge, the accuracy performance drops significantly by 4.53\%, showing out-of-distribution hyperedges have great impact in performance. When the augmented node is added into the same hyperedge with its closest neighbor node from the feature space, the performance of HyperSMOTE slightly drops by 0.25\% in accuracy. HyperSMOTE achieves all-round best performance when the augmented nodes are added to hyperedges with the closest feature.


\section{Conclusion}
We propose HyperSMOTE in this work o alleviate the class imbalance problem on both unimodal and multimodal datasets, which synthesizes node feature based on target nodes from minority classes with their neighbors and trains the decoder on incidence matrix to attach the augmented node into the most relevant hyperedge. Extensive comparison with various baselines on three graph structure unimodal datasets and one multimodal dataset demonstrates that HyperSMOTE is able to augment the minority class with samples that help improve the performance of the model.

\clearpage
\bibliographystyle{IEEEbib}
\bibliography{refs}

\end{document}